\RequirePackage{amsthm}
\documentclass[iicol,sn-mathphys-num,Numbered]{sn-jnl}

\usepackage{graphicx}%
\usepackage{multirow}%
\usepackage{amsmath,amssymb,amsfonts}%
\usepackage{amsthm}%
\usepackage{mathrsfs}%
\usepackage[title]{appendix}%
\usepackage{xcolor}%
\usepackage{textcomp}%
\usepackage{manyfoot}%
\usepackage{booktabs}%
\usepackage{algorithm}%
\usepackage{algorithmic}

\theoremstyle{thmstyleone}%
\theoremstyle{thmstyletwo}%

\theoremstyle{thmstylethree}%

\newcommand{\orcid}[1]{\href{https://orcid.org/#1}{\includegraphics[width=10pt]{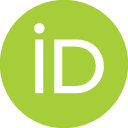}}}

\begin{document}

\title[Article Title]{Grid Jigsaw Representation with CLIP: A New Perspective on Image Clustering}


\author[1]{\fnm{Zijie} \sur{Song\orcid{0000-0002-1262-764X}}}\email{zjsonghfut@gmail.com} 

\author*[1]{\fnm{Zhenzhen} \sur{Hu\orcid{0000-0003-1042-8361}}}\email{huzhen.ice@gmail.com}

\author[1]{\fnm{Richang} \sur{Hong\orcid{0000-0001-5461-3986}}}\email{hongrc.hfut@gmail.com} 

\affil[1]{\orgdiv{the School of Computer Science and Information Engineering}, \orgname{Hefei University of Technology}, \orgaddress{\city{Hefei}, \country{China}}}


\abstract{Unsupervised representation learning for image clustering is essential in computer vision. Although the advancement of visual models has improved image clustering with efficient visual representations, challenges still remain. Firstly, existing features often lack the ability to represent the internal structure of images, hindering the accurate clustering of visually similar images. Secondly, finer-grained semantic labels are often missing, limiting the ability to capture nuanced differences and similarities between images. In this paper, we propose a new perspective on image clustering, the pretrain-based Grid Jigsaw Representation (pGJR). Inspired by human jigsaw puzzle processing, we modify the traditional jigsaw learning to gain a more sequential and incremental understanding of image structure. We also leverage the pretrained CLIP to extract the prior features which can benefit from the enhanced cross-modal representation for richer and more nuanced semantic information and label level differentiation. Our experiments demonstrate that using the pretrained model as a feature extractor can accelerate the convergence of clustering. We append the GJR module to pGJR and observe significant improvements on common-use benchmark datasets. The experimental results highlight the effectiveness of our approach in the clustering task, as evidenced by improvements in the ACC, NMI, and ARI metrics, as well as the super-fast convergence speed.

\textbf{For the official printed version, please visit: \url{https://rdcu.be/d9FkB} \\ 
DOI: \url{https://doi.org/10.1007/s00530-025-01703-x}}
}




\maketitle

\section{Introduction}\label{sec:intr}
Image clustering, as a fundamental task in computer vision, aims to group similar images together based on their visual representations without annotations. As an unsupervised learning task, it revolves around the pivotal task of extracting discriminative image representations. With the advent of deep learning progress, particularly pre-training large-scale vision models in the last two years, researchers have made substantial advancements in image clustering, achieving superior performance compared to traditional methods that relied on handcrafted features~\cite{yang2010image,li2020prototypical,chu2023image,qian2023stable,hoang2024pixel}.

Although deep learning models have revolutionized the field of computer vision by automatically learning hierarchical representations from raw images, there are still limitations in the field of image clustering. First, these supervised learning visual models, i.e., CNN-based~\cite{he2016deep} and Transformer-based~\cite{dosovitskiy2020image}, are trained from the global labels. They primarily focus on differentiating relationships between entire images, while overlooking the internal structure of individuals. However, the internal structural relationships within images hold crucial significance for image representation. Moreover, the annotated labels used in the training process for image classification or object recognition tend to be a single word, which is overly simplistic. Image representations trained on such simple labels only capture the mapping relationship between images and basic labels, failing to provide nuanced discriminative representations. Consequently, directly utilizing image features extracted from these visual models for image clustering tasks remains insufficient.

\begin{figure}[t]
  \includegraphics[width=1.0\linewidth]{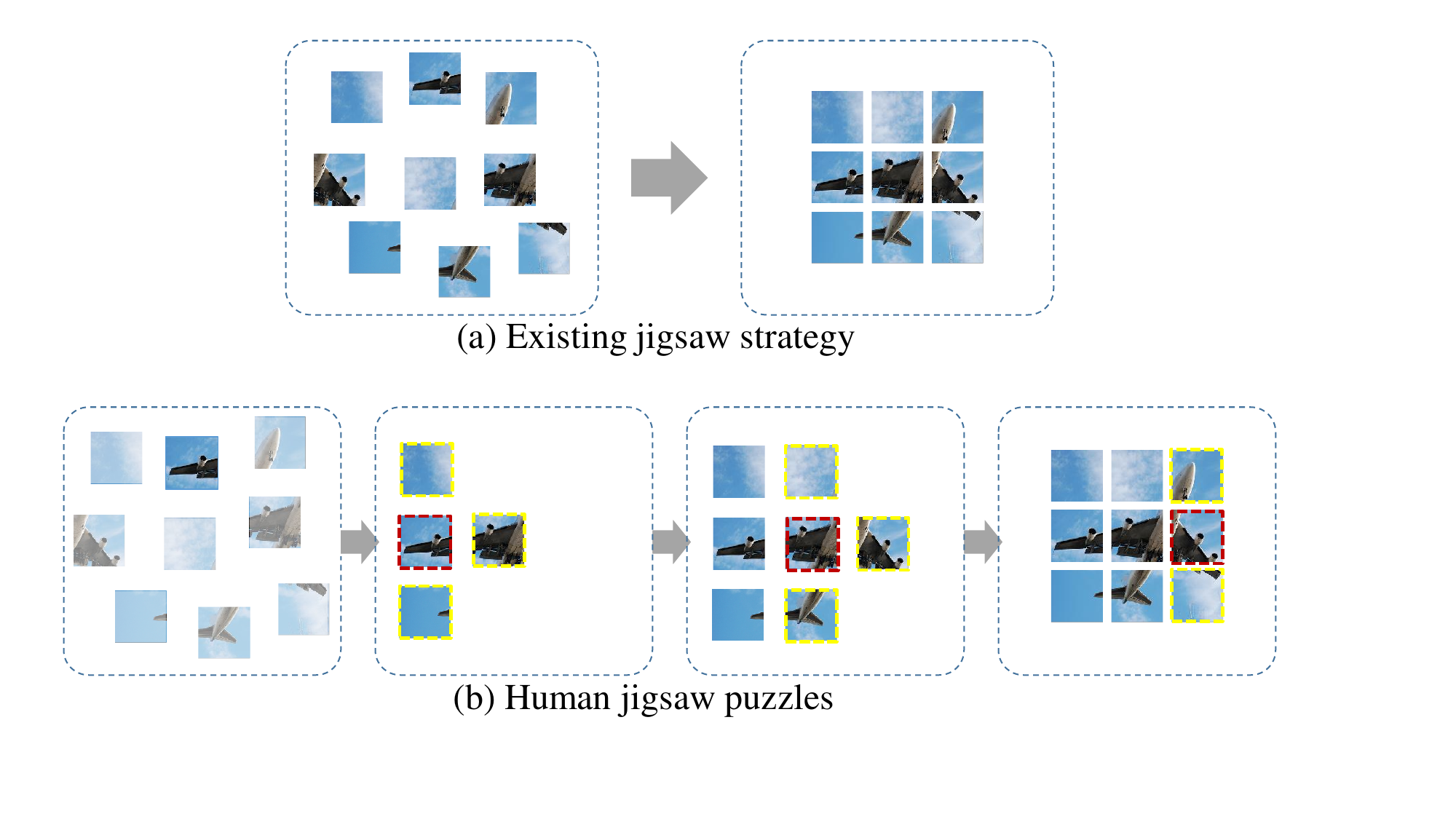}
  \caption{The difference between existing jigsaw strategy and human jigsaw puzzles. Fig.~\ref{fig:1}~(a) presents the existing jigsaw methods which almost focus on learning permutation and sorting with pixels. Human jigsaw puzzles first use one piece as a benchmark to consider the parts around it which is learning splicing and linking by prior image semantics understanding as shown in Fig.~\ref{fig:1}~(b).}
  \label{fig:1}       
\end{figure}

To this end, we address the limitations of existing visual models in the context of image clustering. Self-supervised learning has proven to be an effective approach for learning internal features from data. As a pretext task of self-supervised learning, jigsaw puzzle~\cite{noroozi2016unsupervised} has been shown the ability of exploring the internal structure relationships within images. As shown in Fig.~\ref{fig:1}~(a), by breaking an image into several patches and then reconstructing them, the jigsaw puzzle task aims to capture the internal relationships and spatial dependencies between different regions by shuffling and rearranging all puzzle pieces simultaneously. The achievements from the subsequent researches~\cite{kim2018learning,chen2021jigsaw} demonstrate its potential in uncovering the hidden structural patterns within images. Although Jigsaw puzzle is inspired by human jigsaw solving, the existing Jigsaw puzzle algorithm does not necessarily replicate the exact process of human. In human jigsaw solving, we typically start by identifying a specific puzzle piece and then proceed to locate neighboring pieces around it. This step-by-step approach allows for a gradual construction, focusing on a subset of pieces at a time, as shown in Fig.~\ref{fig:1}~(b). Comparing with the jigsaw puzzle pretext task, the human solving process is a more sequential and incremental understanding of image structure. In our previous work~\cite{song2023grid}, we have preliminarily explored the grid feature based on jigsaw strategy for image clustering and demonstrated its prominent performance via experiments. In this paper, we further elaborate on the breakthrough improvement from pixels in low-level statistics to features on the high-level perception.

In recent years, the integration of vision and language has emerged as a promising research direction in computer vision. Vision and language pre-training models, such as the Contrastive Language-Image Pre-training (CLIP)~\cite{radford2021learning}, utilize large-scale datasets of images and their associated textual descriptions to learn a joint embedding space. By leveraging the joint embedding space provided by CLIP, image clustering algorithms can benefit from the enhanced cross-modal representation for richer and more nuanced semantic information and label level differentiation to foster the development of highly discriminative image representations. In this paper, we replace the convolutional image representation with cross-modal CLIP features to investigate how the cross-modal representation can improve the effectiveness and efficiency of image clustering algorithms. We find that this cross-modal representation not only enhances the accuracy of image clustering but also significantly improves the convergence speed of the clustering algorithms. This acceleration in convergence not only improves the efficiency of image clustering but also facilitates the scalability of the clustering algorithms to larger datasets.

To sum up, we propose a new perspective on image clustering which combines pretrained CLIP visual encoder to extract the prior features and jigsaw strategy to improve clustering performance called pretrain-based Grid Jigsaw Representation (pGJR). Specifically, we first employ a pretrained visual-language model CLIP as a visual extractor to obtain visual representation. Then, we propose the jigsaw supplement method expanded by our previous work GJR~\cite{song2023grid} to fit pretrained representations in training. CLIP pretrained representations provide powerful prior features and GJR as location attention maps module supports more refined adjustment. It is intuitively considered that the nearly finished puzzle with bits of patches in error position or vacancy will not be shuffled again but modify some local positions. We evaluate the effectiveness of our methods for the image clustering benchmarks and provide sufficient ablation study and visualization results.

Our main contributions can be summarized as follows:

\begin{itemize}
\item We propose a new perspective on image clustering which combines pretrained CLIP visual encoder and jigsaw strategy to improve clustering performance named pretrain-based Grid Jigsaw Representation (pGJR) and verify it on the six benchmarks where the results show the great performance on image clustering task.

\item We design a subhuman jigsaw puzzle module to the middle-level visual feature which as a plugin can mine the semantic information on a higher level representation learning. It has strong generalization in both of deep CNN training and combined pretrained model.

\item We exploring the cross-modal representation in the context of image clustering where pretrained CLIP provide mature and learning-friendly representations to improve the performance and efficiency for clustering training.

\end{itemize}

The remainder of this paper is arranged as follows. Sec.~\ref{sec:2} mainly reviews the related work about deep clustering, self-supervised learning and grid feature. Sec.~\ref{sec:GJR} introduces our proposed method named Grid Jigsaw Representation with motivation and algorithm. Sec.~\ref{sec:pGJR} proposes a new perspective on clustering about pre-trained model and jigsaw supplement with pretrain-based visual extractor CLIP, pretrain-based Grid Jigsaw Representation and clustering training process. Sec.~\ref{sec:Exp} presents experimental details, results and ablation study with visualization. Sec.~\ref{sec:con} contains the concluding remarks.

\section{Related Work}
\label{sec:2}

\subsection{Deep Clustering}
Deep clustering~\cite{chang2017deep,wu2019deep,gai2023clustering,zhou2024comprehensive} as a fundamental and essential research direction, mainly leverages the power of deep neural networks to learn high-level features incorporating traditional clustering methods~\cite{ng2002spectral}. The concept of spectral clustering~\cite{shaham2018spectralnet} was introduced to set up input of positive and negative pairs according to calculate their Euclidean distance with classical k-means and promoted many related researches~\cite{bianchi2020spectral,wu2018unsupervised,yang2019deep,tao2021clustering,cai2023learning} to obtain competitive experimental results. 
Li \textit{et al.}~\cite{li2022neural} demonstrated that data augmentation can impose limitations on the identification of manifolds within specific domains, where neural manifold clustering and subspace feature learning embedding should surpass the performance of autoencoder-based deep subspace clustering. Starting with a self-supervised SimCLR~\cite{chen2020simple}, recent visual representation learning methods~\cite{grill2020bootstrap,regatti2021consensus,niu2022spice,li2024contrastive} have achieved great attention for clustering. Tsai \textit{et al.}~\cite{tsai2020mice} leveraged both latent mixture model and contrastive learning to discern different subsets of instances based on their latent semantics. By jointly representation learning and clustering, Do \textit{et al.}~\cite{do2021clustering} proposed a novel framework to provide valuable insights into the intricate patterns at the instance level and served as a clue to extract coarse-grained information in objects.

\subsection{Self-supervised Learning}  
Self-supervised learning has been a thriving field of research for visual representation learning, which aims to extract key semantic information and discriminative visual features from images. One of the pretext tasks involved training the network to reassemble image tiles using jigsaw puzzles~\cite{noroozi2016unsupervised,carlucci2019domain}, establishing a strong knowledge association among the patches of the puzzles. Rather than treating jigsaw puzzles as an independent pretext task, some studies extend this logic of its pattern generalized to more downstream tasks by self-supervised training such as image classification~\cite{dery2017neural,paumard2018jigsaw} and other applications. 
Chen \textit{et al.}~\cite{chen2021jigsaw} introduced a self-supervised learning approach called jigsaw clustering, which involves using disturbed patches as the output and the raw picture as the target for both intra-image and inter-image analysis. Zhang \textit{et al.}~\cite{zhang2022mejigclu} improved the efficiency in contrastive learning with low computational overhead on jigsaw clustering. 

Moreover, typical self-supervised architectures have served as inspiration for representation learning. Wu \textit{et al.}~\cite{wu2018unsupervised} focused on learning feature representations by emphasizing the ability to distinguish individual instances, thus capturing the evident similarities present among instances. Chen \textit{et al.}~\cite{chen2020simple} proposed a simplified contrastive self-supervised learning framework that incorporates learnable nonlinear transformations and effective composition of data augmentations. In this work, Siamese architectures are employed for unsupervised representation learning, with the objective of maximizing the similarity between two augmentations of a single image. 
Bardes \textit{et al.}~\cite{bardes2022vicreg} proposed a variance term that is utilized in both branches of the architecture based on a covariance criterion, which effectively prevents informational collapse and ensures that both branches contribute to the learning process. To address the lack of explicit modeling between visible and masked patches, the context autoencoder~\cite{chen2023context} was proposed to overcome limited representation quality by combining masked representation prediction with masked patch reconstruction.

\subsection{Grid Feature}
The discussion of grid features mainly existed in the object detection task~\cite{he2017mask,ren2015faster} compared with region features about network design and performance. Since Jiang \textit{et al.}~\cite{jiang2020defense} shed light on the potential of grid features on the visual-language task, there has been sparked further interest and exploration in visual representation. Referring to masked words in sentences in natural language processing~\cite{devlin2018bert}, many researches especially those based on transformer structure~\cite{parmar2018image,qi2020imagebert} masked grid units in images to learn the connections between pixels. Huang \textit{et al.}~\cite{huang2020pixel} employed a method where images were directly passed into the feature module pixel by pixel, enabling the learning of local feature relationships in fine detail. This approach allowed for a more granular understanding of the image content. In a similar vein, Dosovitskiy \textit{et al.}~\cite{dosovitskiy2020image} demonstrated that dividing images into 256 patches proved beneficial for vision recognition tasks, which facilitated capturing and processing the image information at a patch level, leading to improved performance in visual recognition tasks. He \textit{et al.}~\cite{he2021masked} designed a powerful approach using masked autoencoders, which involved randomly covering grids of the input image and subsequently reconstructing the missing pixels. Wa \textit{et al.}~\cite{wang2021robust} utilized grid partitioning and decision graphs to efficiently identify clustering centers, thereby enhancing the robustness of the clustering process. However, many of these successfully relied on grid pixels rather than features to conduct large-scale pre-training in order to learn the relationships within the data, and it is worth noting that such approaches often require substantial computing resources.

\section{Grid Jigsaw Representation}
\label{sec:GJR}

\begin{figure*}[t]
  \includegraphics[width=1.0\linewidth]{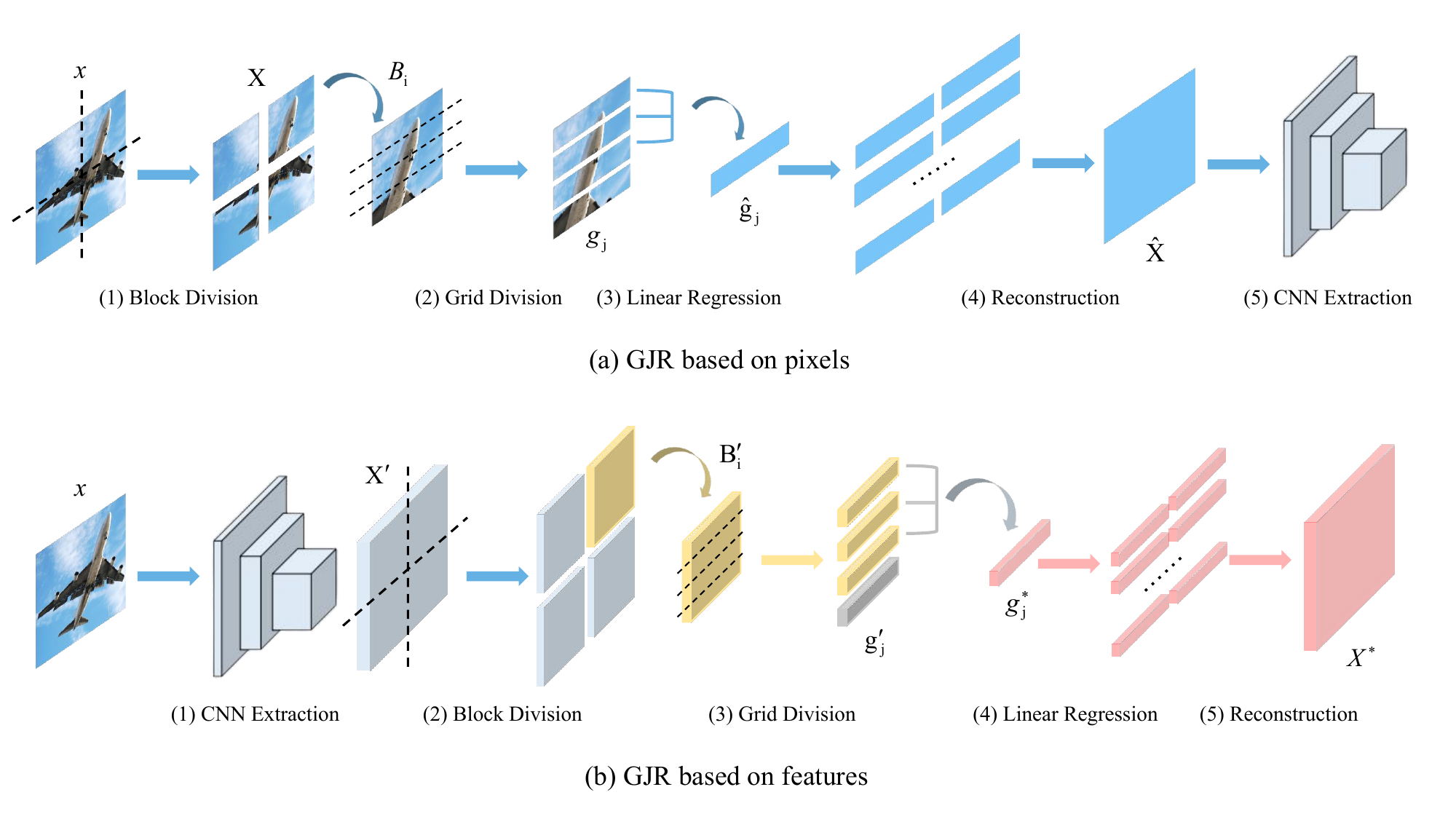}
  \caption{The framework illustration of proposed GJR where Fig.~\ref{fig:2}~(a) is pixel-based version and Fig.~\ref{fig:2}~(b) is feature-based one. There are both five steps about CNN Extraction, Block Division, Grid Division, Linear Regression and Reconstruction, but in different orders.}
  \label{fig:2}       
\end{figure*}

In this section, we introduce the Grid Jigsaw Representation (GJR) methods with two parts: Motivation and Algorithm. It is a complete exposition for GJR which we propose in our previous work~\cite{song2023grid}. The Motivation introduces an improved insight for GJR why we propose a new kind of jigsaw strategy and its conception from pixel to feature (our previous work just preliminarily attempted on grid feature). The Algorithm shows the specific details for GJR where the framework is shown in Fig.~\ref{fig:2}~(b) and the algorithm steps are shown in Alg.~\ref{alg:algorithm}.

\subsection{Motivation}
Unlike supervised learning, unsupervised learning requires a global constraint based on training samples without ground truth labels. In addition to constraints through loss functions~\cite{chen2020simple}, self-supervised learning can also be considered to design diverse network structures according to the characteristics and properties in data or task. This approach aims to abstractly and ingeniously imitate human logic. Jigsaw strategy~\cite{noroozi2016unsupervised} as one of the self-supervised learning methods, mimics human jigsaw puzzles in the analysis, understanding and operation of image patches. 
However, as previously mentioned and illustrated in Fig.~\ref{fig:1}, the notable difference exists between jigsaw strategies and human jigsaw puzzles. The primary distinguishing factor is that most current jigsaw strategies and their extended versions focus on raw image patches, emphasizing low-level statistics such as structural patterns and textures. In the deep neural networks, the feature maps on the top layer of neural networks imply high-level clues for visual representation but may not align with human intuitive perception. Nonetheless, this should not deter us from exploring operations on high-level features, especially since much remains unknown about how the human brain learns. Therefore, we propose to implement the jigsaw strategy on deep-layer features and compare it with pixel-based approaches.

Our previous work~\cite{song2023grid} made a preliminary attempt at developing a jigsaw strategy, which we refer to as Grid Jigsaw Representation (GJR) in this paper. GJR draws inspiration from both jigsaw puzzles and grid features. When presented with pieces of jigsaw puzzles, people naturally tend to infer the position of each piece based on the overall structure of the picture.  Similarly, GJR leverages this intuition and the framework to learn representations by rearranging and reassembling grid patches, enabling the model to capture spatial relationships and contextual information within the image. Jigsaw puzzles inherently imply a certain prior knowledge: the closer the distance between puzzle pieces, the stronger the relevance of the patches will be. When humans solve jigsaw puzzles, they rely on learning cues from the surrounding patches to understand the overall structure of the image, rather than solely relying on the individual patch itself. Taking inspiration from this, we make the assumption that in computer vision, learning visual representation cues from the surrounding feature grids would provide more valuable information than learning from the grid itself. In other words, in a grid feature map separated into blocks, the information from adjacent grids within the same block is more informative than the grid itself. Based on this perspective, we introduce GJR, which replaces a grid with its surrounding grids within the block to facilitate the learning of visual representations. The recent works like BERT~\cite{devlin2018bert} or MAE~\cite{he2022masked} share a similar idea of using masked inputs for prediction during pretraining. However, it is important to emphasize that our GJR differs from these methods because there is no prediction involved as every patch is given. It is more akin to evidence-based association, where the complete image is observed rather than predicting missing parts.

We note that an intuitive presentation may not be enough to capture feature-based advantages or prove that the jigsaw strategy would be better used on features rather than pixels. In this paper, we expand GJR into pixels and demonstrate the advantage of GJR based on features through experimental comparison.

\subsection{Algorithm}
Fig.~\ref{fig:2} shows the framework of GJR respectively based on pixels and based on features. There are five steps in GJR: CNN Extraction, Block Division, Grid Division, Linear Regression and Reconstruction. Notably, the order of the CNN Extraction step differs between the two approaches. In the feature-based GJR, high-level image features are extracted first, followed by the jigsaw operation to obtain new representations. In contrast, the pixel-based GJR extracts features at the end of the process. Since the steps are fundamentally the same, we will focus on introducing the feature-based GJR to detail the specific algorithm.

\begin{algorithm}[t]
\caption{Grid Jigsaw Representation}
\label{alg:algorithm}
\leftline{\textbf{Input}: image $x$}
\leftline{\textbf{Output}: Jigsaw representation $X^*$}
\begin{algorithmic}[1] 
\STATE $X^{'} = CNN(x)$  \\
CNN extraction outputs $n$ feature maps. \\
\STATE $ X^{'}_G = Grid(X^{'}),$  \\
$Grid(\cdot)$ operation as follow: \\
\STATE  $X^{'}_G =\{B^{'}_i\}, i \in [1,m]$  \\
Block division comes to $m$ blocks and every block is $l \times l$ feature. \\
\STATE  $B^{'}_i= \{g^{'}_j\},j \in [1,l]$  \\
Grid division sorts each block by row. \\
\FOR{$i = 1$ to $m$}
\STATE $B^{'}_i=\{g^{'}_j\}, j \in [1,l] $
\FOR{$j = 1$ to l}
\STATE $g^*_j$ = Linear($B^{'}_i$, without $g^{'}_i$)  \\
 Reconstruction of local feature by Linear. \\
\ENDFOR
\STATE $B^*_i = \{ g^*_j \}, j \in [1,l]$
\ENDFOR
\STATE $X^* = \{ B^*_i \}, i \in [1,m]$ 
\STATE \textbf{return} $X^*$
\end{algorithmic}
\end{algorithm}

As shown in Alg.~\ref{alg:algorithm}, given an image $x$ as input, feature maps set $X^{'}$ is extracted by deep Convolutional Neural Network (CNN), which preserves CNN output dimension size $n$:
\begin{equation}
    X^{'} = CNN(x).
\end{equation}

We implemented GJR on feature maps $X^{'}$. Specifically, $X^{'}$ is divided into $m$ blocks and each block has $l \times l$ size, which should meet split size $ n=m \times l \times l $. The objective is to ensure that the grid of image features within each block is appropriately close and relevant to one another. In order to mitigate the computational challenges arising from edge influences and reduce algorithmic complexity, we define row $ g^{'}_j$, $j \in [1,l]  $ as a unit grid for each block $B^{'}_i, i \in [1,m]$. Therefore, we acquire the grid feature maps through a process of region-based and orderly permutation, as defined below:
\begin{equation}
    X^{'}_G = Grid(X^{'}),
\end{equation}
where set $X^{'}_G =\{B{'}_i\}, i \in [1,m]$, set $B{'}_i=\{g^{'}_j\}, j \in [1,l] $. $B{'}_i$ is the block and $g{'}_j$ is the grid. $Grid(\cdot)$ just implements the division operation and has not changed $X^{'}$, so $X^{'}_G$ keeps its value. In this way, we can assume that every unit grid in each block has a strong semantic correlation because of its close distance. We extract and integrate other grids except $g^{'}_j$ in $B_i$ to reconstruct $g^*_j$. $g^*_j$ as the unit grid of jigsaw representation has the same size as $g^{'}_j$ with linear regression:
\begin{equation}
    g^*_j = ReLU(\omega_j (\sum_{g^{'}\ne g^{'}_j}^{B_i} g^{'}) + \delta_j),
\end{equation}
where $\omega_j$ is the trainable weight and $\delta_j$ is the vector bias. Then the new representation $X^*$ is reconstructed by all $g^*_j$. The representation $X^{'}$ with $B^{'}_i$ as block, $g^{'}_j$ as unit grid is transformed into the jigsaw representation $X^*_i$ with $B^*_i$ as block, $g^*_j$ as grid. Note that our reconstruction operates during the forward propagation stage and does not involve predicting a specific target, which is mainly apart from other methods. Equally important, the practical significance of $B^*_i$ is totally different from $B^{'}_i$. $X^*$ holds a higher dimension which integrates the relationship of information in the graphic area. Each grid feature in $B^*_i$ is a collection of adjacent information of the original grid feature in $B^{'}_i$.

\section{Pre-trained Model and Jigsaw Supplement}
\label{sec:pGJR}

\begin{figure}[t]
  \includegraphics[width=1.0\linewidth]{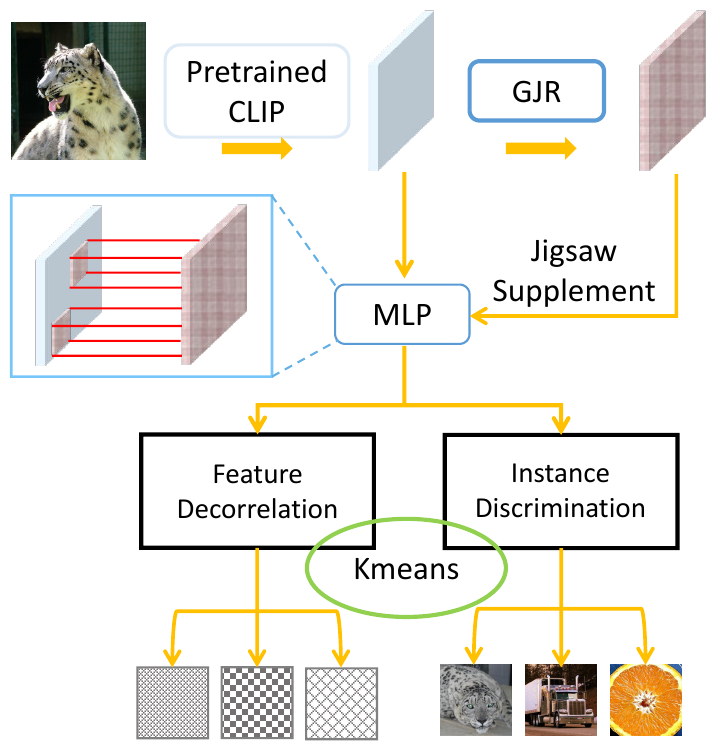}
  \caption{The framework illustration of proposed pGJR. pGJR is consisted of CLIP-based representation structure and jigsaw supplement for image clustering. The representation learning includes jigsaw feature network in forward prorogation and clustering constraint optimization in backward propagation.}
  \label{fig:3}       
\end{figure}

In this section, we introduce the new perspective on image clustering where pretrained image-text model CLIP~\cite{radford2021learning} is used as a visual extractor to replace CNN for training. This pattern can efficiently improve the convergence speed of clustering. As a pretrained image-text encoder, CLIP employs highly differentiated text-guided visual modeling during training to generate more reasonable and meaningful image representations. Moreover, we provide pretrain-based Grid Jigsaw Representation (pGJR) which can be a supplement to further improve the clustering performance in such a pattern. Finally, we introduce the training process for image clustering. The framework is shown in Fig.~\ref{fig:3}.

\subsection{Pretrain-based Visual Extractor CLIP}

We employ a pretrained visual-language model CLIP~\cite{radford2021learning} as a visual extractor to replace the CNN extractor in Sec.~\ref{sec:GJR}. It utilizes the pretrained representations from the CLIP visual encoder as prior features. We first propose CLIP(tuned) as our new baseline, which provides better and faster clustering representations.

Given an image x as input, representation $X_C$ is extracted by pretrained CLIP. The output here inherits the dimension 768 from the CLIP final layer:
\begin{equation}
    X_C = CLIP(x).
\end{equation}

Next, a multi-layer perceptron (MLP) is employed to transform features from 768 dimensions to n, where n is set to match the final dimension of GJR's $X^*$ for clustering. In this context, the MLP functions as a simple linear layer, effectively serving as a linear transformation:
\begin{equation}
    \hat{X_C} = Linear(\textbf{W}_C X_C + \Delta_C),
\end{equation}
where $\textbf{W}_C$ is the trainable weight matrix and $\Delta_C$ is the vector bias group.
Subsequent experiments will prove pretrained representation $\hat{X_C}$ as initial value with powerful prior information which provides efficient convergence speed and improves the baseline of image clustering. We find that it is already state-of-the-art on most of our test dataset, but it also shows some limitations in fine-grained ability,  which will be discussed later in Sec.~\ref{sec:Exp}.

\subsection{Pretrain-based Grid Jigsaw Representation}

Moreover, we expand our GJR methods with CLIP feature called pGJR. Get CLIP representation $X_C$ first. Then, pGJR handles $X_C$, as the same as the operation in Alg.~\ref{alg:algorithm} to obtain $X^*_C$:
\begin{equation}
    X^*_C = GJR(X_C).
\end{equation}

It is same to CLIP(tuned) that the representation should be transformed from 768 to n through a MLP. But there is little different and not direct use $X^*_C$:
\begin{equation}
    \hat{{X^*_C}} = Linear^*(\textbf{W}^*_C [X_C + ReLU(X^*_C)] + \Delta^*_C),
\end{equation}
where $\textbf{W}^*_C$ is the trainable weight matrix and $\Delta^*_C$ is the vector bias group. $ReLU(X^*_C)$ will be not a reconstruction representation but a region attention to replenish $X_C$. It is considered that CLIP as a mature visual feature extractor provides powerful prior information to generate representation. Thus, it is no use to global reconstruction to disrupt again. Turning to jigsaw puzzles, people will not shuffle afresh the jigsaw when it comes to finish with bits of patches in error position. Thus, only the local features need to be strengthened or modified. Finally, we use $\hat{{X^*_C}}$ as pGJR output for clustering.

\subsection{Clustering}
The clustering task necessitates the ability to distinguish objects based on the representation of the images themselves. It should not only focus on the relationships between adjacent regions within a single sample but also effectively differentiate similar features across different samples. We believe that GJR presents a promising method for visual representation in clustering tasks. Its motivation is rooted in learning the splicing and linking of visual semantics through the jigsaw strategy.

We apply the representation learning method IDFD~\cite{tao2021clustering} with simple k-means to obtain the clustering results, where the Instance Discrimination~\cite{wu2018unsupervised} is utilized to capture the similarity between instances and Feature Decorrelation~\cite{tao2021clustering} is utilized to reduce correlations within features. Both GJR and pGJR incorporate this module following the visual representation features. Given an unlabeled dataset $\{x_i\}^n_{i=i}$, every image $x_i$ is handled and reduced dimension by a fully connected layer to obtain Jigsaw representation $X^*_i$. Then, we define the whole representation set $V = \{v_i\}^n_{i=1} = \{X^*_i \}^n_{i=i}$ to be set with a predefined number of clusters $k$.

Given $x_i$ corresponding representation $v_i$, Instance Discrimination controls data $x_i$ classified into the $i$th class. The $v_i$ as weight vector can be calculated with the probability of $v$ being assigned into the $i$th class:
\begin{equation}
    P(i|v) = \frac{exp(v^T_i v_i/\tau_1)}{\sum^n_{j=1}{exp(v^T_j v_i/\tau_1})},
    \label{eq4}
\end{equation}
where $v^T_i v_i$ is to evaluate how match degree $v_i$ with the $j$th class, and $\tau_1$ is the temperature parameter.
Then the objective function $L_I$ of instance discrimination is as follows:
\begin{equation}
     L_I = -\sum^n_i{log(\frac{exp(v^T_i v_i/\tau_1)}{\sum^n_{j=1}{exp(v^T_j v_i/\tau_1})})}.
\end{equation}

Feature Decorrelation imposes constraints on features between different images and fits GJR in the backward propagation stage. It defines latent vectors $F = V^T = \{f_l\}^d_{l=1}$. Unlike Eq.~\eqref{eq4}, the new constraint is transformed to:

\begin{equation}
    Q(l|f) = \frac{exp(f^T_l f_l/\tau_2)}{\sum^d_{m=1}{exp(f^T_m f_l/\tau_2})},
\end{equation}
where $Q(l|f)$ is similar to $P(i|v)$ but the implication of the transposed feature $f$ will be completely different semantic information from $v$. $\tau_2$ is another temperature parameter. The objective function $L_D$ of feature decorrelation is as follows:

\begin{equation}
     L_D = -\sum^m_l{log(\frac{exp(f^T_l f_l/\tau_2)}{\sum^d_{m=1}{exp(f^T_m f_l/\tau_2})})}.
\end{equation}

To sum up the two calculation results of $L_I$ and $L_D$, the whole objective function is shown as:
\begin{equation}
     L = L_I + L_D.
\end{equation}

\section{Experiments}
\label{sec:Exp}

In this section, we first introduce the datasets and evaluation metrics. Then, we show and analyze the main results for GJR and pGJR respectively. By contrast, representation ability of GJR is obvious, and convergence efficiency for pGJR is emphasized. Finally, ablation study demonstrates generalizability performance and shows the feature representations distribution.

\subsection{Datasets and Metrics}

\begin{table}[t]
  \centering
	\renewcommand{\arraystretch}{1.1}
  \begin{tabular}{llll} \hline 
    Dataset  & Images & Clusters & Size \\ 
    \hline
    STL-10 &13,000 & 10  & 96 × 96 × 3 \\
    ImageNet-10 &13,000 & 10  & 96 × 96 × 3 \\
    CIFAR-10 &50,000 &10 & 32 × 32 × 3 \\
    ImageNetDog-15 & 19,500 & 15 & 96 × 96 × 3 \\
    CIFAR-100/20 &50,000 & 20  & 32 × 32 × 3 \\
    ImageNetTiny-200 & 100,000  & 200 & 64 × 64 × 3 \\ 
    \hline
  \end{tabular}
  \caption{Statistics of different datasets.}
  \label{tab:1}       
\end{table}

Following the six common used benchmarks, we conduct unsupervised clustering experiments on STL-10~\cite{coates2011analysis}, ImageNet-10~\cite{deng2009imagenet}, CIFAR-10~\cite{krizhevsky2009learning}, ImageNetDog-15~\cite{deng2009imagenet}, CIFAR-100/20~\cite{krizhevsky2009learning} and ImageNetTiny-200~\cite{deng2009imagenet}. We summarize the statistics and key details of each dataset in Table~\ref{tab:1} where we list the numbers of images, number of clusters, and image sizes of these datasets. Specifically, the training and testing sets of dataset STL-10 were jointly used and images from the three ImageNet subsets were resized as shown. We follow the three metrics: standard clustering accuracy~(ACC)~\cite{li2006relationships}, normalized mutual information~(NMI)~\cite{strehl2002cluster}, and adjusted rand index~(ARI)~\cite{hubert1985comparing}. ACC measures the proportion of samples that are correctly classified in a clustering result, out of the total number of samples. NMI is an information-theoretic metric used to evaluate the similarity between the clustering results and the ground truth class labels. ARI is a corrected-for-chance version of the Rand Index, which measures the similarity between the clustering results and the ground truth class labels.
The higher the percentage of these three metrics, the more accurate clustering assignments. Every experiment result can be trained on two NVIDIA GeForce RTX 3060 for GJR and one enough for pGJR.

\subsection{Implementation Details}

\begin{table}[t]
  \centering
	\renewcommand{\arraystretch}{1.1}
    \resizebox{0.48\textwidth}{!}{
  \begin{tabular}{lllll} \hline 
    Datasets   &  $\eta_{0}$ &  Epoch(gamma) & Block & Grid \\
    \hline
    STL-10 & 3e-2 & 800/1200(0.1) & 2 & 8 × 8 \\
    ImageNet-10 & 3e-2 & ---  & 2 & 8 × 8 \\
    CIFAR-10 & 2e-2 & 800/1300/1800(0.1) & 8 & 4 × 4  \\
    ImageNetDog-15 & 3e-2 & 600/950/1300/1650(0.1) & 8 & 4 × 4  \\
    CIFAR-100/20 & 3e-2 & 800/1800(0.1) & 2  & 8 × 8  \\
    ImageNetTiny-200 &  3e-2 & 600/950/1300/1650(0.1) & 8  & 4 × 4 \\
    \hline
  \end{tabular}
  }
  \caption{Hyperparameter setting for GJR of different datasets. $\eta_{0}$ is initial learning rate. Epoch(gamma) shows which epoch to reduce learning rate and its ratio.}
  \label{tab:2}       
\end{table}

For the GJR results, we adopted the best comprehensive effect ResNet18~\cite{he2016deep} as the basic neural network architecture for GJR and easy reproduced clustering method strategy with IDFD~\cite{tao2021clustering} and kmeans. 
Our experimental settings and data augmentation strategies are just in accordance with IDFD. The total number of epochs was set to 2000, and the batch size was set to 128. Output feature dimension size was set to $ n=128 $ from ResNet18.
Temperature parameters are set as $\tau_1 = 1$ and $\tau_2 = 2$. The parameters in GJR module, such as block number $m$ and grid number $l$, are set according to the size of feature maps with specific deep CNN. The size of grid tensor $n$ will be certain after the two values product of $m$ and $l$ are determined $n=m \times l \times l$. For example, $n = 128$ when $m = 8$ and $l = 4$. Our main hyperparameters for GJR are grid numbers to control $m$ and $l$ and learning rate to control training rhythm as shown in Table~\ref{tab:2}. 

\begin{table}[t]
  \centering
	\renewcommand{\arraystretch}{1.1}
    \resizebox{0.48\textwidth}{!}{
  \begin{tabular}{lllll} \hline 
    Datasets  &  $\eta_{0}$ &  $\eta_{1}$(epoch)  & Block &  Grid  \\
    \hline
    STL-10 & 2e-2  & 2e-3(100)  & 2 & 8 × 8 \\
    ImageNet-10 & 5e-2  & 1e-3(30)  & 8 & 4 × 4 \\
    CIFAR-10 & 5e-1  & 1e-2(75) & 8 & 4 × 4 \\
    ImageNetDog-15 & 5e-2  & 5e-3(45) & 2 & 8 × 8 \\
    CIFAR-100/20 & 2e-2  & 2e-3(75) & 2  & 8 × 8 \\
    ImageNetTiny-200 & 5e-2  & 5e-3(100) & 8  & 4 × 4 \\
    \hline
  \end{tabular}
  }
  \caption{Hyperparameter setting for pGJR of different datasets. $\eta_{0}$ is initial learning rate. $\eta_{1}$(epoch) is learning rate of second stage and its epoch.}
  \label{tab:3}       
\end{table}

For the pGJR results, we maintain the clustering strategy using IDFD and k-means. Regarding CLIP(tuned), the backbone ResNet18 is replaced by the pretrained CLIP~\cite{radford2021learning} for image feature extraction, and a linear layer is added as the MLP. Specifically, this means that the CLIP visual extractor is frozen, and only the parameters of the linear layer are trained. The experimental settings and data augmentation strategies remain the same as those used in GJR, so we will not repeat them. The only change is the substitution of the linear layer with the GJR module, as illustrated in Alg.\ref{alg:algorithm} for pGJR. Hyperparameters for pGJR are adjusted to enhance training performance, as detailed in Table~\ref{tab:3}.

\subsection{Main Results}

\begin{figure}[t]
  \includegraphics[width=1.0\linewidth]{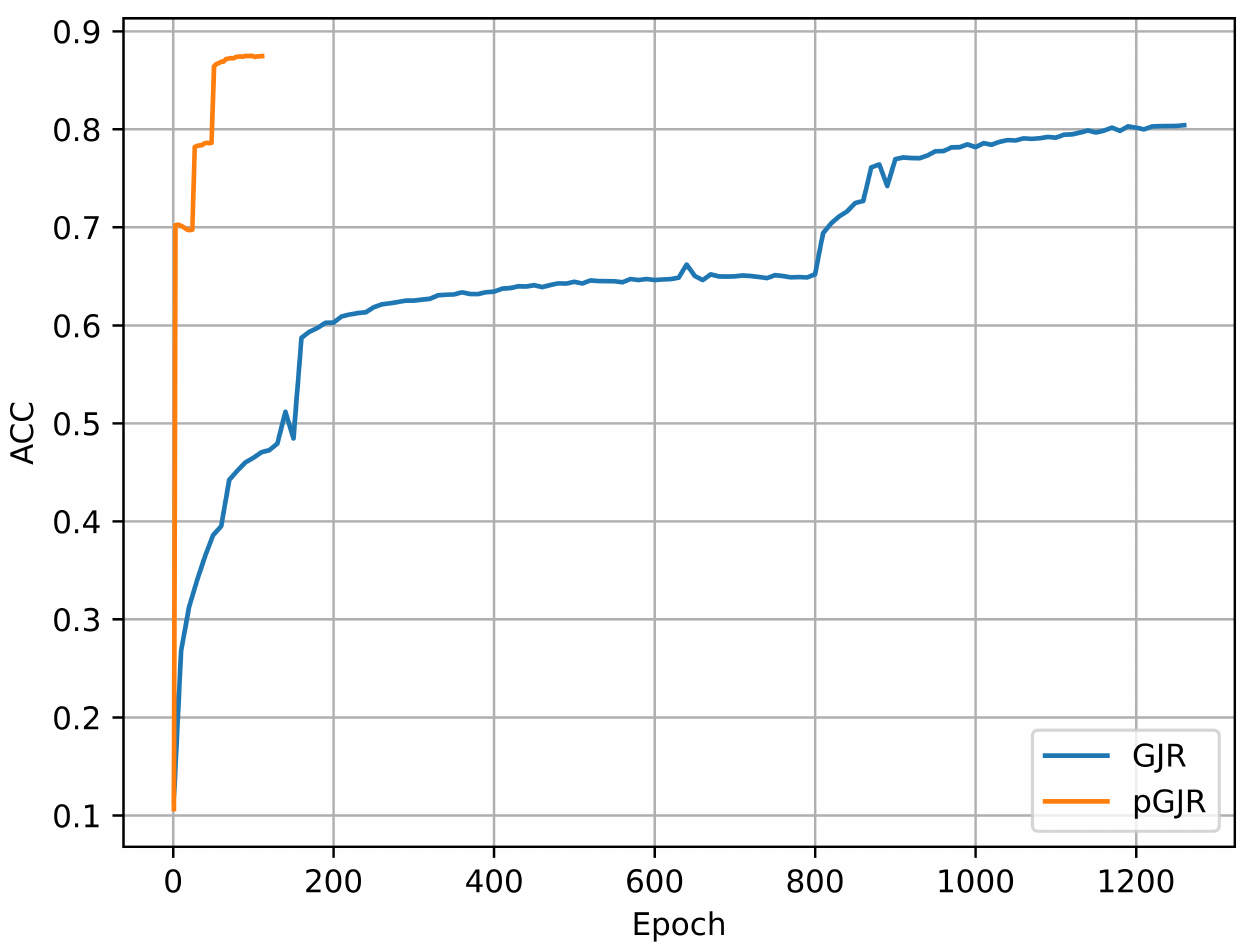}
  \caption{Convergence efficiency and performance compared with GJR and pGJR on CIFAR-10.}
  \label{fig:4}       
\end{figure}

We first compare the convergence efficiency between GJR and pGJR in Fig. ~\ref{fig:4}, which intuitively illustrates the rationale behind our work expansion. It can be observed that GJR with the initial ResNet18 requires nearly 1,200 epochs to begin converging and approximately 2,000 epochs to approach optimal performance in our experiments. In contrast, pGJR with the pretrained CLIP model reaches and exceeds this performance level in just 150 epochs. The training time per epoch is largely similar for both GJR and pGJR. More epoch comparisons have been shown in Table~\ref{tab:2} and Table~\ref{tab:3}. Additionally, pretrained models like CLIP are relatively efficient in terms of GPU memory usage and can be utilized at no cost. It's important to note that clustering tasks do not necessarily require large models with the highest performance, as unsupervised learning for unlabeled samples still relies on effective unsupervised algorithms for training.

\begin{table*}[t]
  \centering
	\renewcommand{\arraystretch}{1.3}
    \resizebox{1.0\textwidth}{!}{
  \begin{tabular}{llllllllllllllll} \hline 
    Datasets  & \multicolumn{3}{l}{STL-10} & \multicolumn{3}{l}{ImageNet-10} & \multicolumn{3}{l}{CIFAR-10} & \multicolumn{3}{l}{ImageNetDog-15}  & \multicolumn{3}{l}{CIFAR-100/20}\\ 
    \hline
    Metric(\%)  & ACC  & NMI  & ARI & ACC  & NMI  & ARI & ACC  & NMI  & ARI & ACC  & NMI  & ARI & ACC &  NMI &  ARI \\
    \hline
     SCAN~\cite{van2020scan}  & \textcolor{lightgray}{80.9}  & \textcolor{lightgray}{69.8}  & \textcolor{lightgray}{64.6} & ------ & ------ &  ------  & \textcolor{lightgray}{88.3} &  \textcolor{lightgray}{79.7} &  \textcolor{lightgray}{77.2} & ------  & ------ &------\ & \textcolor{lightgray}{50.7}  & \textcolor{lightgray}{48.6}  & \textcolor{lightgray}{33.3} \\
    IMC-SwAV~\cite{ntelemis2022information} & \textcolor{lightgray}{85.3} &  \textcolor{lightgray}{74.7}  & \textcolor{lightgray}{71.6} & ------ & ------ & ------ & \textcolor{lightgray}{89.7}   & \textcolor{lightgray}{81.8}  & \textcolor{lightgray}{80.0} & ------ &  ------ & ------  & \textcolor{lightgray}{51.9}  & \textcolor{lightgray}{52.7} &  \textcolor{lightgray}{36.1} \\
    TCL~\cite{li2022twin} & \textcolor{lightgray}{79.9} &  \textcolor{lightgray}{86.8} &  \textcolor{lightgray}{75.7} & \textcolor{lightgray}{87.5}  & \textcolor{lightgray}{89.5}  & \textcolor{lightgray}{83.7} & \textcolor{lightgray}{81.9}  & \textcolor{lightgray}{88.7}  & \textcolor{lightgray}{78.0} & \textcolor{lightgray}{67.5}  &  \textcolor{lightgray}{62.7}  &  \textcolor{lightgray}{52.6} & \textcolor{lightgray}{53.8}  & \textcolor{lightgray}{56.7}  & \textcolor{lightgray}{38.7} \\
    SPICE~\cite{niu2022spice} & \textcolor{lightgray}{93.8} &  \textcolor{lightgray}{87.2} &  \textcolor{lightgray}{87.0} & \textcolor{lightgray}{95.9}  & \textcolor{lightgray}{90.2}  & \textcolor{lightgray}{91.2} & \textcolor{lightgray}{92.6}  & \textcolor{lightgray}{86.5}  & \textcolor{lightgray}{85.2} & \textcolor{lightgray}{67.5}  &  \textcolor{lightgray}{62.7}  &  \textcolor{lightgray}{52.6} & \textcolor{lightgray}{53.8} &  \textcolor{lightgray}{56.7}  & \textcolor{lightgray}{38.7} \\
    \hline
    DEC~\cite{xie2016unsupervised} &35.9  & 27.6 &  18.6 &38.1 &  28.2 &  20.3 &30.1  & 25.7  & 16.1 & 19.5 &   12.2 & 7.9  &18.5  &   13.6 & 5.0 \\
    DAC~\cite{chang2017deep} &47.0 &  36.6 &  25.7 &52.7 &  39.4 &  30.2 &52.2  & 39.6  & 30.6 &27.5  & 21.9  & 11.1 &23.8  & 18.5 &  8.8 \\
    DCCM~\cite{wu2019deep} &48.2  & 37.6  & 26.2 &71.0 &  60.8  & 55.5 &62.3  & 49.6  & 40.8 &38.3 &  32.1  & 18.2 & 32.7  & 28.5  & 17.3  \\
    IIC~\cite{ji2019invariant} &59.6  & 49.6  & 39.7 &------ & ------ &------ &61.7  & 51.1 &  41.1 &------  &------& ------ & 25.7 &  22.5 &  11.7\\
    PICA~\cite{huang2020deep} &71.3  & 61.1 &  53.1 &87.0  & 80.2 & 76.1 &69.6 &  59.1 &  51.2 &35.2  & 35.2 & 20.1 & 33.7 &  31.0 & 17.1\\
    DRC~\cite{zhong2020deep} &74.7 &  64.4 &  56.9 &88.4  & 83.0 &  79.8 &72.7 &  62.1 &  54.7 &38.9 &  38.4  &23.3 & 36.7   &35.6  & 20.8 \\
    MiCE~\cite{tsai2020mice} &72.0 &  61.3 &  53.2 &------ & ------ &------ &83.5 &  73.7 &  69.5 &39.0  & 39.0 & 24.7 & 42.2 &  43.0  & 27.7 \\
    CC~\cite{li2021contrastive} & 85.0 &  74.6  & 72.6 & 89.3 &  85.9 & 82.2 & 79.0 &  70.5  & 63.7 &42.9  & 43.1 &  26.6 & 42.9  & 44.5 &  27.4 \\
    IDFD~\cite{tao2021clustering} &75.6  & 64.3 &  57.5 &95.4  & 89.8  & 90.1 & 81.5 &  71.1  & 66.3 & 59.1  & 54.6 &  41.3 & 42.0 &  42.6  & 26.4 \\
    CRLC~\cite{do2021clustering} &81.8  & 72.9 &  68.2 &85.4 &  83.1 & 75.9 &79.9  & 67.9  & 63.4 &46.1  & 48.4 & 29.7 & 42.5   &41.6  & 26.3 \\
    NNCC~\cite{xu2022deep} &72.5 & 61.6 & ------  &75.1 & 68.3 & ------  &81.9 &  73.7 & ------ & 43.8  & 42.1 & ------ &40.1   & 37.2 & ------ \\
    NMCE~\cite{li2022neural} &72.5  & 61.4  & 55.2 &90.6  & 81.9 & 80.8 &83.0 &  76.1  & 71.0  &39.8 &  39.3 & 22.7 & 43.7  & 48.8  & 32.2 \\
    SRL~\cite{wang2023structure} &72.4 &  81.8 & 68.2 &90.2  & \underline{95.9}  & 91.2 & 83.4 & \textbf{90.3} & \textbf{81.7} & 42.3  & 27.8 &  14.6 & 50.7  & 51.6 &  34.2\\
    ICC-SPC~\cite{guo2023improving} &77.7 &  69.5 &  64.1 &95.2 &  89.2 & 89.8 &83.2  & 78.4  & 75.2 & 47.4  & 49.3 &  32.2 & \textbf{64.1} & \textbf{60.8} & \textbf{50.4} \\
    JDCE~\cite{liang2024jointly} & 87.0 & 77.3 & 74.7 & 94.4 & 89.7 & 90.2 & 84.3 & 71.1 & 68.8 & \underline{59.9} & 55.6 & 40.3 & 47.2 & 41.8 & 32.5 \\
    \hline
    GJR(pixels) &71.4 &  60.0 &  51.9 & 87.5 &  79.0  & 75.6 & 79.8  & 69.8  & 63.8 & 36.8 &  37.3 &  22.3 & 44.2 &  41.0   &25.7\\ 
    GJR(features) &78.2 & 68.9  & 59.6 &96.2 &  91.3 &  91.9 & 83.7  & 75.0  & 70.2 & \textbf{63.7}& \textbf{61.0} & \textbf{47.0}  & 46.1 &  45.9  & 29.6 \\ 
    \hline
    CLIP(tuned) & \underline{97.6}  & \underline{94.1}  & \underline{94.7} &\underline{98.2}  & 95.6  & \underline{96.0} & \underline{87.0} &  79.0  & 71.5 & 53.6  & 54.5  & 42.3 & 57.2 &  56.2 &  38.1 \\
    pGJR & \textbf{97.9} & \textbf{94.7} & \textbf{95.3} & \textbf{98.8} & \textbf{96.9} & \textbf{97.4} & \textbf{87.5}   &\underline{79.3}  & \underline{72.6} & 57.8  & \underline{56.8} &  \underline{44.3}  & \underline{58.3}  & \underline{58.2}   & \underline{38.2}\\
    \hline
  \end{tabular}
  }
  \caption{Evaluating clustering results (\%) on five datasets compared with state-of-the-art. Our performance is trained with pretrained frozen model CLIP. The best results are highlighted in \textbf{blod} and the second are highlighted by \underline{underline}. References to \textcolor{lightgray}{grey} use pseudo-tags or semi-supervised algorithms.}
  \label{tab:4}       
\end{table*}

Table~\ref{tab:4} shows GJR and pGJR performance with recent advanced clustering methods which are mainly listed by published year. GJR compared with other advanced clustering methods in our previous work~\cite{song2023grid} has obtained high performance in main clustering methods. Our method focuses on representation learning relying only on ResNet18 architecture and kmeans. We also test our GJR based on pixels, although we think it may not make sense from the beginning design, and the experiment result also proves it. We emphasize GJR mimics human jigsaw logic which requires prior processing features rather than low-level pixels.

Our proposed results by CLIP(tuned) and pGJR all just use kmeans for unsupervised clustering and train few linear layers parameters within low training cost. Thus, the SOTA methods about SCAN~\cite{van2020scan}, IMC-SwAV~\cite{ntelemis2022information}, TCL~\cite{li2022twin} using pseudo-tags which are listed on the paper-with-code and SPICE~\cite{niu2022spice} adopting semi-supervised algorithms are set in gray and are not compared in Table~\ref{tab:4}. It can be found that our proposed method mostly obtained the best and second performance. Even our pGJR are SOTA methods and exceed semi-supervised algorithms SPICE~\cite{niu2022spice} on STL-10, ImageNet-10 and CIFAR-100/20. Compared with traditional methods, it is not surprising that our pretrained strategy has improved distinctly. Then, introduced GJR module makes further achievement from CLIP(tuned) which obtains the best or the second results. However, we find that the pretrained strategy does not perform well on the fine-grained dataset ImageNetDog-15. Considering CLIP training, the matching pair associated with dogs is simply 'The photo of a dog' which fails to capture the subtle differences among various dog breeds. As a result, the representations provided by the pretrained CLIP model are unsatisfactory for ImageNetDog-15. Although our proposed pGJR method improves NMI and ARI even with short-stage training by effectively distinguishing between samples, we still recommend using GJR based on ResNet for training on fine-grained datasets instead of relying on pretrained CLIP.

\begin{table}[t]
  \centering
	\renewcommand{\arraystretch}{1.1}
  \begin{tabular}{llll} 
    \hline
    Method &ACC &NMI &ARI  \\
    \hline
    PICA~\cite{huang2020deep} & 9.8  & 27.7 & 4.0 \\
    CC~\cite{li2021contrastive} & 14.0  & 34.0 & 7.1\\
    NNCC~\cite{xu2022deep} & 14.1  &  33.3 & 16.1\\
    SPICE~\cite{niu2022spice}  & 30.5  & 44.9 & 16.1 \\
    IMC-SwAV~\cite{ntelemis2022information} & 27.6  &  49.8 & 14.5\\
    SRL~\cite{wang2023structure} & 27.8 &  42.3 & 14.6 \\
    \hline
    pGJR  & \textbf{43.1}  & \textbf{58.1}  &\textbf{28.7}\\
    \hline
  \end{tabular}
  \caption{Evaluating clustering results (\%) on ImageNetTiny-200 compared with state-of-the-art.}
  \label{tab:5}       
\end{table}

Considering large scale and clustering categories of the ImageNetTiny-200, the cost of training on this dataset is much larger and there are relatively few comparable clustering results as shown in Table~\ref{tab:5}. In unsupervised learning, it is difficult to learn large-scale types from feature difference and Euclidean distance. However, our pGJR also uses only 150 epochs and can obtain state-of-the-art performance by tuning.

\subsection{Ablation Study and Visualization}

\begin{table}[t]
  \centering
	\renewcommand{\arraystretch}{1.1}
    \resizebox{0.48\textwidth}{!}{
  \begin{tabular}{llllllll} 
    \hline
    \multirow{2}{*}{Dataset}  &Backbone &\multicolumn{3}{l}{ResNet18} &\multicolumn{3}{l}{Pretrained CLIP}\\
                          &Metric  & ACC  & NMI  & ARI   & ACC  & NMI  & ARI \\
    \hline
    \multirow{3}{*}{CIFAR-10} &Net(tuned)  & 95.4  & 89.8  & 90.1   & 98.2  & 95.6  & 96.0 \\
                          &Net + GJR & 96.2  & 91.3  & 91.9  & 98.8  & 96.9  & 97.4\\
                          &Up(+)  &\textcolor{blue}{0.8}   & \textcolor{blue}{1.5} 
   & \textcolor{blue}{1.8} &\textcolor{blue}{0.6}  &  \textcolor{blue}{1.3}   & \textcolor{blue}{1.4}\\
    \hline
    \multirow{3}{*}{ImageTiny-200} &Net(tuned)  & 9.3  & 25.0  & 3.2  & 42.4  & 57.6  & 28.0 \\
                          &Net + GJR & 9.4   & 25.5  &  3.4  & 43.1  & 58.1  & 28.7\\
                          &Up(+)  &\textcolor{blue}{0.1}   & \textcolor{blue}{0.5} 
   & \textcolor{blue}{0.2}  &\textcolor{blue}{0.7}  & \textcolor{blue}{0.5}   & \textcolor{blue}{0.7}\\
    \hline
  \end{tabular}
  }
  \caption{Clustering results (\%) for GJR module on ImageNet-10 and ImageNetTiny-200. Net(tuned) means only backbone as framework. Up(+) means the improvement provided by GJR module colored in \textcolor{blue}{blue}.}
  \label{tab:6}       
\end{table}

Firstly, it is underlying to embody performance for GJR module, as shown in Table~\ref{tab:6}. We maintain each parameter according to the specified settings. Although pGJR uses the pretrained CLIP to extract the feature without training the backbone, we think this serves as another indication of the applicability of GJR. We choose ImageNet-10 as the small-scale dataset and ImageNetTiny-200 as the large-scale one where clustering training on ImageNetTiny-200 is significantly more challenging than on ImageNet-10. While ResNet18 appears to have limited capacity for large-scale datasets, increasing its depth incurs substantial training costs. Thus, pretrained model should be considered for image clustering training on large-scale datasets, where CLIP(tuned) and pGJR can be a suitable baseline.

\begin{figure}[t]
  \includegraphics[width=1.0\linewidth]{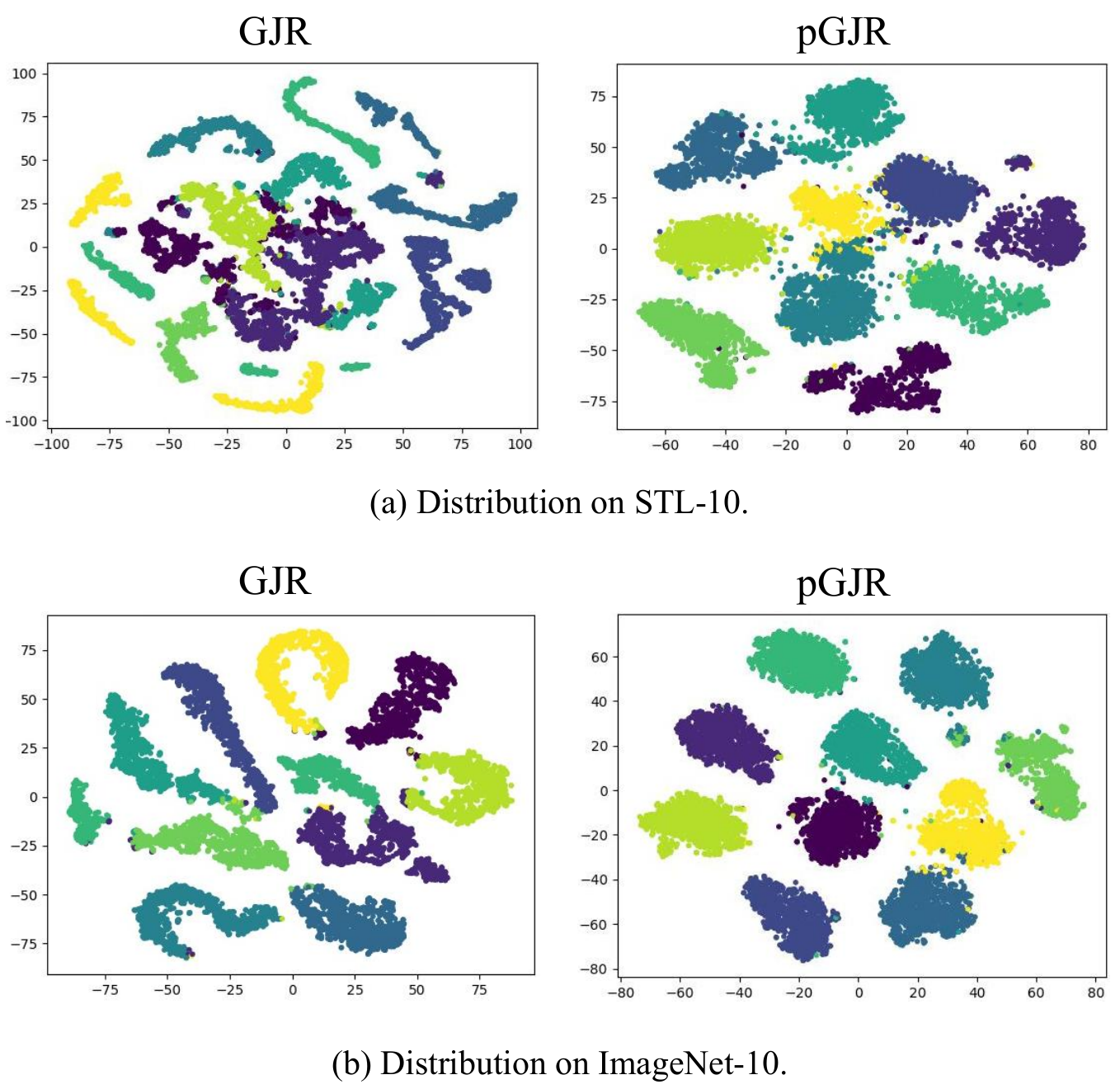}
  \caption{Visualization of clustering feature representations distribution compared with GJR and pGJR.}
  \label{fig:5}       
\end{figure}

Then, we show the clustering feature representations distribution compared with GJR and pGJR in Fig.~\ref{fig:5} on STL-10 and ImageNet-10. The distributions show the preference for clustering between GJR and pGJR. It is clearly that every kind of clusters is compact and narrow for GJR which determines cleaner boundaries and distances between distinguishable categories due to long-period training. However, instances of the same species may be spread across multiple areas, as seen with yellow for GJR on STL-10 in Fig.~\ref{fig:5}~(a) left. pGJR shows more evenly distributed clusters with the highest metrics. There are sufficiently distinct contours and obvious clustering centers with a few hard samples misclassified.

\begin{figure*}[t]
  \includegraphics[width=1.0\linewidth]{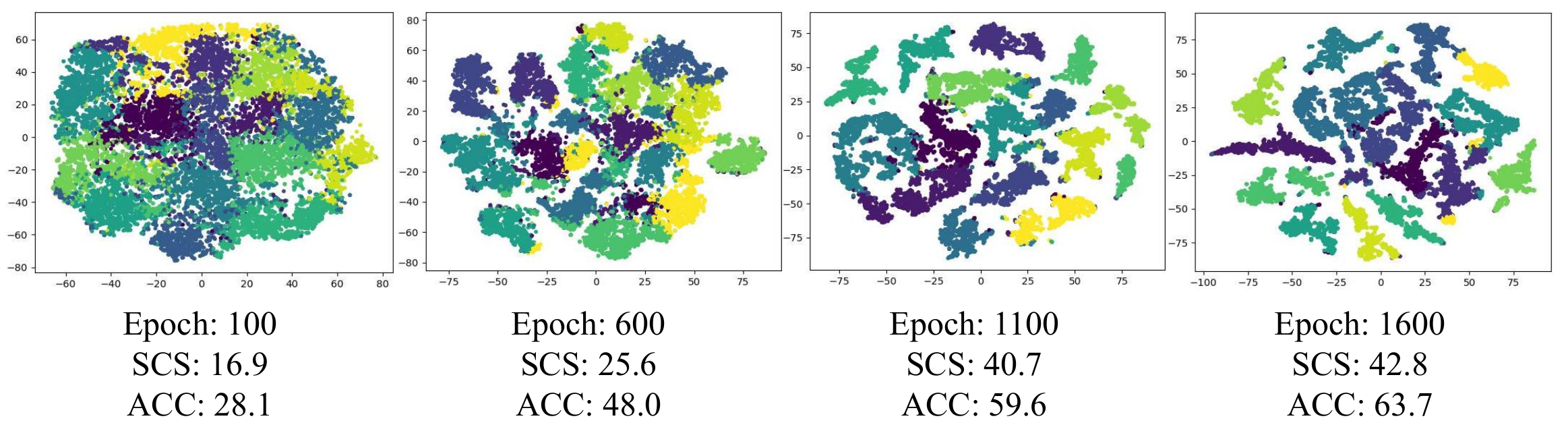}
  \caption{Visualization of clustering feature representations distribution compared with GJR and pGJR on ImageNetDog-15.}
  \label{fig:6} 
  \vspace{-10pt}
\end{figure*}

Considered the specificity on fine-gained dataset, we show the visualization distribution for GJR on ImageNetDog-15 in Fig.~\ref{fig:6}. We print the figures every 500 epochs from 100 to 1600. This dataset has 15 categories for dogs, so there are more centers to train and cluster. Compared to pGJR, more epochs are required to process. However, the steady training process and the increasingly tight clusters symbolize the effectiveness of GJR. Here, we provide evaluative Silhouette Coefficient Score (SCS)~\cite{aranganayagi2007clustering} which presents the contour of clusters. Both of distributions and SCS demonstrate the effectiveness of clustering centers with training samples aggregation.

\begin{figure*}[t]
  \includegraphics[width=1.0\linewidth]{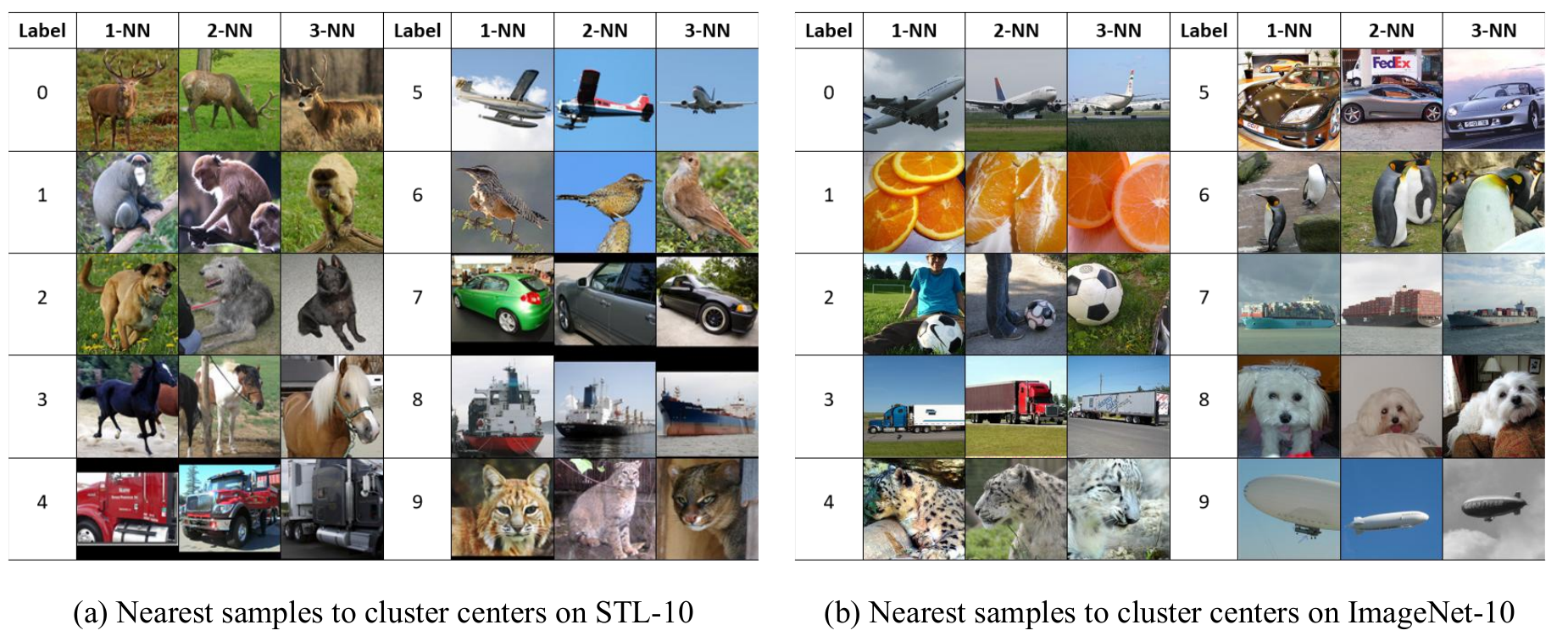}
  \caption{Visualization of semantic clusters on STL10 in Fig.~\ref{fig:7}~(a) and ImageNet-10 in Fig.~\ref{fig:7}~(b). The top three nearest samples called K-NearestNeighbor (KNN) to the cluster centers are shown with 1-NN, 2-NN and 3-NN.}
  \label{fig:7}       
\end{figure*}

Finally, we analyze semantic clusters through visualization of K-Nearest-Neighbor (KNN) and we set $k=3$. Fig.~\ref{fig:7} shows the top three nearest samples of the cluster centers which we find by calculating the Euclidean distance between the samples and their respective cluster centers. It proved that the nearest samples exactly match the human annotations and gather in their discriminative regions with the cluster centers. For example, the cluster with label ‘0’ captures the ‘deer’ class on STL-10, and its most discriminative regions capture the planes at different locations. Moreover, the cluster with label ‘4’ captures the ‘leopard’ class on ImageNet-10 where the 1-NN and 2-NN samples have the same motion and perspective with just little difference in shade of color and background.

\section{Conclusion}
\label{sec:con}

In this paper, we introduce a new perspective on image clustering through jigsaw feature representation (GJR) and a pretrained visual extractor. Specifically, we expand GJR into a more flexible module that initially applies the jigsaw strategy to grid features. We systematically explain the motivation and design behind this approach, highlighting the discrepancies between human and computer perception, from pixel to feature. Furthermore, we propose a novel method of using the pretrained model CLIP as a feature extractor, which can accelerate the convergence of clustering. We also innovate the pretrained Grid Jigsaw Representation (pGJR) by integrating our GJR with CLIP to enhance clustering performance. The experimental results demonstrate the effectiveness of our methods in visual representation learning and training efficiency for unsupervised image clustering. Additionally, we compare GJR and pGJR, particularly on fine-grained datasets, to provide guidance on the appropriate use of pretrained models in various contexts.

\backmatter






\bibliography{GJR}

\end{document}